\pdfoutput=1

\documentclass[11pt]{article}
\usepackage{setspace}
\usepackage[dvipsnames]{xcolor}
\usepackage{graphicx}
\usepackage{tabularx}
\usepackage{afterpage}
\usepackage{booktabs}
\usepackage{amssymb,amsmath}
\usepackage{subcaption}
\usepackage{xspace}
\usepackage{etoolbox}
\usepackage{enumitem} %
\usepackage{cellspace} %
\usepackage{acl}
\providetoggle{showcomments}
\toggletrue{showcomments}

\usepackage{times}
\usepackage{latexsym}
\usepackage{xspace}
\usepackage[T1]{fontenc}

\usepackage[utf8]{inputenc}

\usepackage{microtype}

\usepackage{inconsolata}
\usepackage{float}

\title{Hanging in the Balance: \\ Pivotal Moments in Crisis Counseling Conversations}

\newcommand{\nsone}{\hspace*{.1in}}
\author{\nsone\nsone Vivian Nguyen \nsone Lillian Lee \nsone Cristian Danescu-Niculescu-Mizil\thanks{Senior corresponding author.} \\
Cornell University \\ \texttt{vn72@cornell.edu} \nsone \texttt{llee@cs.cornell.edu} \nsone \texttt{cristian@cs.cornell.edu}}

\newcommand{\cut}[1]{}
\newcommand{\xhdr}[1]{{\noindent\bfseries #1.}}

\newcommand{\range}{range\xspace}
\newcommand{\Range}{Range\xspace}
\newcommand{\pivotal}{pivotal\xspace}
\newcommand{\highp}{high-pivotal\xspace}
\newcommand{\lowp}{low-pivotal\xspace}
\newcommand{\Pivotal}{Pivotal\xspace}
\newcommand{\PIV}{\mathrm{PIV}}
\newcommand{\disengagement}{disengagement\xspace}

\definecolor{therapist}{RGB}{36,97,189}
\definecolor{client}{RGB}{244,130,130}

\begin{document}
\maketitle

\begin{abstract}

During a conversation, there can come certain moments where its outcome hangs in the balance.
In these \emph{\pivotal} moments, how one responds can put the conversation on substantially different trajectories leading to 
significantly 
different outcomes.
Systems that can detect when such moments arise could assist conversationalists in 
domains with highly consequential outcomes,
such as mental health crisis counseling.

In this work, we introduce an unsupervised computational method for  detecting such \pivotal moments as they happen.
The intuition is
that a moment is \pivotal if our expectation of the conversation's \textit{outcome} varies widely depending on what might be said next. 
By applying our method to crisis counseling conversations, 
we first validate it by showing that it aligns with human perception---counselors take significantly longer to respond during moments detected by our method---and with the eventual conversational trajectory---which is more likely to change course at these times.
We then use our framework to explore the relation between the counselor's response during \pivotal moments and the eventual outcome of the session.

\end{abstract}

\section{Introduction}
\label{sec:intro}
During a conversation, some moments are especially consequential: how one responds in these moments can put the conversation on substantially different trajectories leading to significantly different outcomes.
Such \emph{\pivotal} moments are especially relevant in high-stakes domains with challenging conversations, such as mental health crisis counseling \cite{pisani_individuals_2022}.
These are the  moments in which conversational skills 
are most needed, as the outcome of the counseling sessions hangs in the balance.
In itself, making sure that counselors don't miss such moments is an important challenge in crisis counseling. %

\begin{figure}[t]
    \centering
    \includegraphics[width=0.95\linewidth]{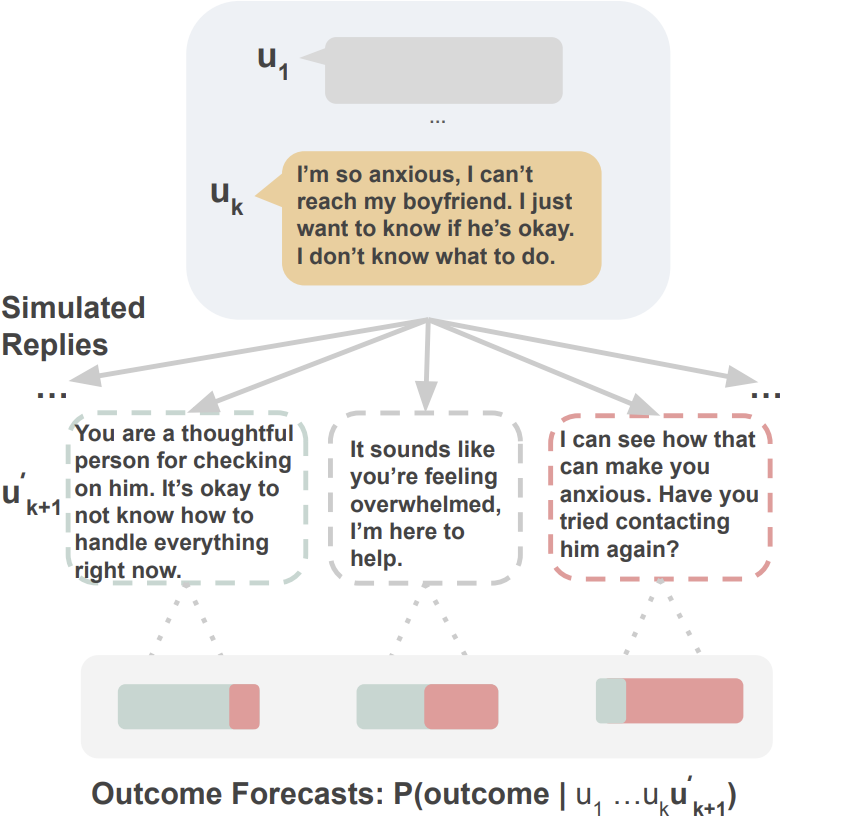}
    \caption{
    A \textit{pivotal} moment: the \textit{next} (yet to be produced) response $u_{k+1}'$ is \textit{expected} to have a large impact on the conversation's eventual outcome (i.e., its probability  varies widely based on the response).
}
    \label{fig:intro_figure}
\end{figure}

Automatically detecting \pivotal moments as they happen could greatly aid the counseling process.
During the session, it could be used to signal the counselor to pay particular attention to their response, or to identify moments when a supervisor should step in \cite{gallant_bug---ear_1989}.
After the session, these moments could be used to help counselors and supervisors retrospectively analyze the session and discuss alternative responses \cite{de_jong_effect_2014,wang_practice_2024_fixed,hsu_helping_2023}. 

Identifying such moments automatically, however, presents several major challenges.
First, no labels are available, and collecting such labels would be especially difficult in high stake domains like crisis counseling, due to the level of experience required for identifying them and to the privacy concerns surrounding the data.
Second, these moments reflect counterfactual possibilities that we can never observe in real data.
Thus, we face 
a radically unsupervised problem.

In response, we propose a formulation that draws from the broader concept of suspense from econometrics \cite{ely_suspense_2015}, 
and that
that we  adapt to the conversational domain.
Consider the analogy of chess: we can say that a moment is suspenseful if the choice of \textit{next} move (e.g., white advances the pawn or the rook) is \textit{expected} to have a large impact on who wins the game.
A spectator 
can imagine the space of likely next moves and intuitively estimate how each might affect the outcome based on the mechanics of the game:
if these estimates have high variance, they feel suspense.
Importantly, this concept is \textit{ex ante}, referring to the moment before the next (yet to be executed) move.\footnote{Suspense contrasts with \textit{surprise}, which is a retrospective (\textit{ex post}) concept referring to a change in expectations that results from an already executed move.}

Our proposed formalization for \pivotal moments adapts 
the concept of suspense
to the conversational domain: 
a moment in a conversation is \emph{pivotal} if the \textit{next} response is \textit{expected} to have a large impact on the conversation's final outcome (see \autoref{fig:intro_figure}).
However, in conversations, the space of all possible next ``moves'' is much broader: most utterances can accept infinitely many responses.
Additionally, conversational mechanics are much more complex, making it much harder to estimate how a response might impact the eventual outcome of the conversation.
We address these challenges through recent NLP advances: we use large language models to sample from the space of likely responses ($u_{k+1}'$'s in Figure \ref{fig:intro_figure}) and we rely on conversational forecasting to estimate how these responses will impact the final outcome of the conversation (by estimating $P(\mathrm{outcome}|u_1...u_k\mathbf{u_{k+1}'})$ in \autoref{fig:intro_figure}).

In partnership with one of the largest text-based crisis counseling services in the US, we apply this methodology to identify \pivotal moments in counseling conversations.
We focus on an outcome that is particularly consequential in this setting: whether the person in crisis disengages and abandons the counseling session before it is over \cite{cox_how_2021,gould_crisis_2022}. 
Considering the lack of labels for \pivotal moments, we must rely on extrinsic validation of our measure.
In particular, we show that in moments that our method infers to be \pivotal, counselors spend more time composing a response, indicating that they perceive it as important.
A retrospective analysis additionally shows that the actual counselor response in \pivotal moments is more likely to change the expected trajectory of the conversation.

We also use our framework to conduct an exploratory analysis of the relation to the eventual outcome of the counseling session.
We find that more than at any other times during an eventually-successful session, in \pivotal moments counselors are better able to improve the trajectory.
The reverse is true for eventually-unsuccessful sessions: they are most likely to degrade during \pivotal moments.
Thus, the way counselors tilt the balance at \pivotal moments strongly correlates with the eventual outcome of the conversation.

Finally, we conduct a qualitative analysis to characterize  moments that are identified as being \pivotal by our framework. 
We find that such moments are more likely to occur when the person in crisis expresses uncertainty about how to proceed, when they explicitly ask for advice, or when they self-disclose something significant.

In summary, in this work we:
\begin{itemize}
    \item introduce a formalism defining \pivotal moments and a methodology for detecting them in an online fashion;
    \item apply and validate our methodology in a crisis counseling setting;
    \item analyze the relation between counselor responses in \pivotal moments and the eventual outcome of the counseling sessions.
\end{itemize}

While this work focuses on the crisis counseling domain, our \pivotal-moments framework is general.  
Future work could apply it to other high-stake conversational domains such as education or political debates, or to improve human-AI interactions.  
To encourage such new applications, we release the code together with a demonstration on public (non-sensitive) online conversations as part of ConvoKit.\footnote{\url{https://convokit.cornell.edu/}}

\section{Crisis Counseling Setting}
\label{sec:data}
We develop our methodology in the context of Crisis Text Line, a free, 24/7 crisis counseling service that provides support for individuals in mental health crises \cite{althoff_large-scale_2016,gould_crisis_2022}.
The individuals in crisis, henceforth \textit{texters}, engage in conversations with counselors on the platform via text-based messaging.
One of the authors went through the counselor training program to familiarize themselves with the process and challenges faced by the counselors.
In this work, we accessed data from January 2015 to October 2020, containing over $1.5$ million conversations, after redaction of personally-identifiable information by the platform.\footnote{This study was approved by Cornell's IRB.}
Messages are timestamped, allowing us to analyze the timing of interactions within the conversations.

In general, the counselor's overall objective in each conversation is to guide the texter towards a calmer state.
However, during a session, the texter may abandon the conversation partway through.
A key challenge of online counseling is avoiding texter \textit{\disengagement}, or lack of conversational closure \cite{cox_how_2021}.
When compared to in-person counseling, the online environment makes it particularly easy for the texters to abandon the session, and particularly hard for the counselor to do anything about it, leading to situations that are particularly distressing.

In this work, we focus on identifying \pivotal moments with respect to the outcome of texter \disengagement.
In a conversation, counselors follow a standardized protocol to close the conversation when faced with an unresponsive texter.
Thus, to determine if a session results in \disengagement, we check if the counselor's last message in the conversation contains key phrases such as ``stepped away from your phone,'' ``haven't heard from you [in a while],'' ``wanted to check in,'' or ends with unanswered questions containing ``?''. 
We consider conversations in which the texter disengages as \emph{unsuccessful}.
In contrast, we consider a \emph{successful} outcome as one where the texter indicates in a survey following their session, that the counselor was helpful.
In this scenario, the texter successfully concludes their conversation with a sense of support and closure.

In our overall analysis, we consider a dataset of $1,000$ such conversations, half of which ended with \disengagement, the other half successfully, paired by the length of the conversation.
We further truncate each conversation by the last three turns to exclude explicit signals of the session's ending. 
We also use a separate dataset of $5,000$ paired conversations for training the forecaster model and another one of $10,000$ conversations for fine-tuning the language model (Section \ref{sec:method}).

\section{Formalizing \Pivotal Moments}
\label{sec:method}
Intuitively, a conversation reaches a \pivotal moment when what is said next matters.
Here, we explore two possible approaches formalizing this intuition and two corresponding unsupervised methods to identify them in an online manner: one that considers the \emph{range} of possible replies at each given moment and one that relies on the more general concept of \emph{``suspense''} from econometrics. 

\subsection{Formalization}

\xhdr{Baseline formalization: \Range} A straightforward attempt to formalize the intuition behind \pivotal moments is to consider how broad the \textit{\range} of possible replies is at that moment.
This approach generalizes the concept of open versus closed questions, where open questions leave room for more variation in the possible replies than closed ones, and thus open more directions the conversation can take \cite{dohrenwend1965effects,pomerantz2005linguistic,alic_computationally_2022}.
Similarly, an utterance in a conversation (whether a question or not) that allows for more varied responses---and thus more ways in which the conversation can develop---could be considered as being more potentially impactful.
To calculate this \Range at a given moment $k$, we can first simulate possible responses using a large language model, and then calculate the average cosine distance between the vector representations of each such simulated response and the mean of all their vector representations. 

However, this formalization has a major conceptual limitation.
A large \Range value does not necessarily imply that choosing between responses is consequential.
Responses that appear to be semantically different may still lead to similar eventual outcomes. 
For example, a question such as ``What is your name?'' elicits a wide range of possible responses, as different individuals would provide different names, but the impact on the conversation is typically minimal.

\xhdr{Our formalization: ``suspenseful'' moments}
We propose a formalism that addresses the conceptual limitation of the baseline approach discussed above: instead of considering the range of possible \emph{replies}, we consider the variation of the \emph{expected outcomes} that these possible replies yield.
This formalism draws from the more general concept of ``suspense'' from econometrics,
where suspenseful moments are defined as those in which the variance of the next moment's outcome expectations are greater  \cite{ely_suspense_2015}. 
Recall the chess analogy given in Section ~\ref{sec:intro}.
Our formalization of \pivotal moments adapts the concept of suspense to the conversational domain, where next responses correspond to next moves in the chess analogy.
Importantly, this concept is \textit{ex ante}, referring to the moment before the next (yet to be composed) response.

\subsection{Operationalization}
Two key elements of this formalization are particularly challenging to operationalize in the conversational domain.
First the space of likely responses is extremely broad. 
Here, like with the baseline \Range approach, we can use large language models to generate a sample of likely responses.

Second, conversational mechanics are complex and nondeterministic, making estimation of the impact of a possible response on the conversation's outcome especially challenging.
Here we propose the use of \textit{conversational forecasting}, which is the task of predicting the eventual outcome of an ongoing conversation as it develops, based on the current conversational context \cite{chang_trouble_2019,kementchedjhieva_dynamic_2021,altarawneh_conversation_2023}.
Using a conversational forecasting model, we can compute how the likelihood of the outcome 
varies according to the possible responses.

More concretely, to quantify how pivotal moment $k$ is in an ongoing conversation, we first simulate $n$ likely responses at this moment: $\{u_{k+1}'^1, u_{k+1}'^2, ..., u_{k+1}'^n\}$.
For each generated response $u_{k+1}'$, we use a forecasting model to predict the likelihood of the conversation's outcome, considering the prior context and this hypothetical response: $P(\mathrm{outcome} | u_1u_2...u_k\mathbf{u_{k+1}'})$. 
This likelihood represents the predicted probability of the conversation's outcome following each hypothetical path with $u_{k+1}'$ as the next utterance.
We can then measure how pivotal a moment at timestep $k$ is as the variance in these probabilities across all the different paths:
\begin{equation*}
{\PIV}_k = {\mathrm{Var}}_{u_{k+1}'}[P(\mathrm{outcome} | u_1u_2...u_k\mathbf{u_{k+1}'})]\, .
\end{equation*}
\Pivotal moments occur when $\PIV_k$ is high (large variance).
This means that the outcome of the conversation is highly sensitive to the choice of response at that particular moment. 
In other words, our expectation of the outcome varies widely depending on what could be said next. 
Conversely, low $\PIV_k$ (small variance) characterizes non-\pivotal moments, as the conversation's outcome is not as sensitive to the specific choice of response at this point. 
In our analysis, we consider (1) the continuous measure $\PIV_k$ and (2)~a discretization based on the top 
10th percentile 
and bottom 10th percentile to compare \highp and \lowp moments.

\subsection{Implementation}\label{sec:implementation}

Our method relies on two main components: a \textit{simulator} model for generating possible responses and a \textit{forecaster} model for forecasting the eventual outcome of the conversation.
\textit{Note that this work was done on secure internal servers due to the private nature of the data; we are therefore limited in the size of the models we can employ.} 
We propose a general framework for identifying \pivotal moments, such that the particular implementations of the components (e.g., models used for simulation or for forecasting) can and should be changed according to the application domain and to resource availability.
We release our code in a modular fashion, making the change of these components straightforward.

\xhdr{Simulator}
For our simulator model, we used Llama-3.1-8B \cite{grattafiori_llama_2024}, fine-tuned on a dataset of $10,000$ conversations. 
We employed a $90/10$ split for training and validation, and trained for $2$ epochs, with LoRA rank = $16$, context length $2048$, batch size $2$, learning rate $\mbox{2e-4}$, and AdamW optimizer \cite{loshchilov_decoupled_2017}, achieving a validation perplexity of $5.59$.
To generate potential counselor responses in the context of the conversation, we set the temperature to $0.8$ and used multinomial sampling as the decoding strategy to achieve an adequate diversity in the generations. 
For our measure, we generate $n=10$ possible continuations at each texter message in the conversation and limit the response length to a maximum of $60$ tokens.

\xhdr{Forecaster}
We also train a separate model to forecast the likelihood of the texter eventually disengaging (Section \ref{sec:data}).
We apply this model at every moment $k$ to produce $P(\mathrm{disengagement} | u_1u_2...u_k\mathbf{u_{k+1}'})$ for each simulated continuation $u_{k+1}'$.
Following prior work on conversational forecasting \cite{kementchedjhieva_dynamic_2021}, we fine-tuned RoBERTa-large \cite{liu_roberta_2019} using a dataset of $5000$ conversations, half of which ended successfully 
(i.e., conversations after which the 
texter rated their counselor as helpful) and the other half with the texter disengaging partway through the conversation.
We created 
pairs of successful/unsuccessful conversations matched on length (number of utterances); each conversation was truncated by the last three 
utterances
to remove explicit signals of the conversation's ending.  
All conversations used for training are discarded in the analysis below. 
We trained for $5$ epochs, with learning rate $\mbox{1e-5}$, batch size $16$, and AdamW optimizer \cite{loshchilov_decoupled_2017}.
We experimented with modifying the model along various implementation axes.\footnote{We tried combinations of \{BERT, RoBERTa\} $\times$ \{static training, dynamic training \cite{kementchedjhieva_dynamic_2021}\}  $\times$  \{conversation segment chunking, no chunking\}.   What we report is for the best set of choices: RoBERTa, static, no chunking.}
The model achieves a forecasting accuracy of $73\%$, following the evaluation methodology from prior work on conversation forecasting \cite{chang_trouble_2019}.

While the forecasting performance is not perfect, it outperforms results from other domains,
e.g., forecasting conversation derailment \cite{chang_trouble_2019,kementchedjhieva_dynamic_2021,altarawneh_conversation_2023}. Even with far-from-perfect performance, forecasting was shown to be effective in real-world applications \cite{chang_thread_2022}.

\section{Application to Crisis Counseling}
\label{sec:results}
We now apply our method to identify and examine \pivotal moments in crisis counseling conversations.
Given that counselors are expected to guide the conversation, we focus our analysis only on the moments in which the counselor (rather than the texter) is about to respond.
This renders any insights from this analysis more interpretable and actionable since the platform can more easily implement counselor-facing interventions.

We start with an extrinsic validation of $\PIV_k$, showing that it aligns with our expectations about counselor behavior and conversational trajectory. 
We then use our framework to analyze the effects of responses at \pivotal moments. 

\subsection{Extrinsic validation}

\xhdr{Response time} 
As \pivotal moments reflect points in the conversation where what is about to be said matters, we would expect counselors to carefully consider their responses in these moments. 
To test whether our formulation captures this intuition, we analyze the actual response times of counselors following moments our method identifies as high- vs. \lowp.\footnote{We note that response time was not used at any point of our pivotalness measuring pipeline.} 
As shown in \autoref{tab:response_time}, 
we find that 
counselors take significantly longer to reply in \highp moments (on average they take $7.5$ seconds longer, $p<0.001$, according to a Mann-Whitney U test; the difference between the 75th percentiles is $18$ seconds). 
This observation aligns with our intuition that \pivotal moments may require more thought and deliberation, in consideration of how subsequent responses could significantly shape the conversation's trajectory.
In addition to serving as an extrinsic validation of our framework, this difference suggests that \pivotal moments might be moments when assistance (e.g., from human supervisors) would be most needed, although future work would be necessary to explore 
the potential of such an application.

Notably, this result is not confounded by the length of the response: there is no significant difference in the length of counselor replies in high-$\PIV$ moments vs. low-$\PIV$ ones ($p=0.17$; Mann-Whitney U test).
Additionally, the naive Range formulation does not capture this intuitive difference: counselors take similar time to reply regardless of how varied the space of likely responses is ($p=0.266$).
This highlights the importance of the forecasting component of our framework, which allows us to focus on the variation of expected outcomes, rather than of just possible responses.
\begin{table}
\centering
\begin{tabular}{llll|l}
\hline
\textbf{Measure} & \textbf{High} & \textbf{Low} & \textbf{Diff} & \text{\bf p-Value} \\
\text{$\PIV$} & \text{102.03} & \text{94.53} & \text{7.50} & 0.001\textcolor{red}{*} \\
\text{\Range} & \text{90.35} & \text{88.36} & \text{1.99} & 0.266 \\
\hline
\end{tabular}
\caption{
Counselors take longer on average to respond during \highp moments (in seconds) compared to \lowp moments as identified by our measure ($\PIV$).  The baseline \Range formulation does not reflect this difference.
}
\label{tab:response_time}
\end{table}

\xhdr{Retrospective trajectory improvement} 
\begin{figure*}[t]
    \centering
    \includegraphics[width=1\linewidth]{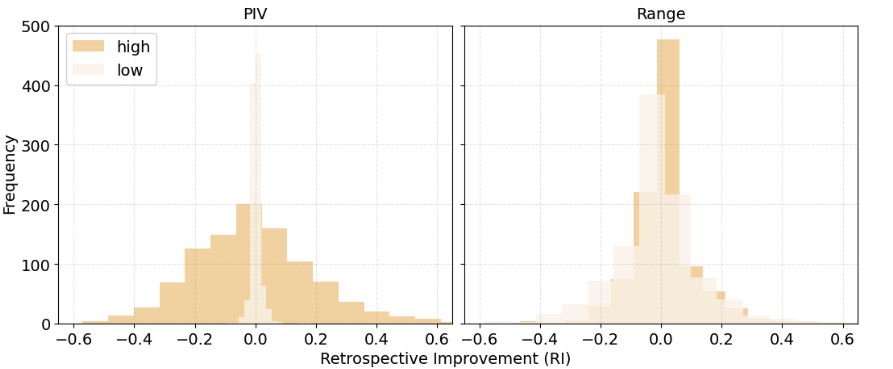}
    \caption{(Left) Counselor responses in \highp moments can greatly improve (positive x-axis) or degrade (negative x-axis) the trajectory, while responses in \lowp moments leave the outcome probability largely unchanged. (Right) This difference is not apparent when using the baseline \Range formulation. 
    }
    \label{fig:overall_critical_noncritical_surprise}
\end{figure*}
\Pivotal moments are characterized by the \textit{potential} for the conversation's trajectory to change in a consequential way.
To check if this potential is actually realized in moments our framework identifies as \pivotal, we do a retrospective analysis of the conversations to check what the counselor eventually responded in these moments.
It is important to note that our method detects \pivotal moments \textit{ex ante}, without access to the actual counselor response. 
Therefore, we can validate our measure by using the hindsight knowledge of how the conversations actually unfolded.

To do so, we consider a retrospective measure of \textit{trajectory improvement}, the degree to which our expectation of the outcome improves from before to after counselor's reply.
We can calculate the trajectory improvement at moment $k$ as 
the change in predicted probability that the texter will eventually disengage from moment $k$ to moment $k+2$:
\begin{align*}
RI_{@k} = P(\textrm{\disengagement}|u_1...u_{k}) - \\  P(\textrm{\disengagement} | u_1...u_{k+2})
\end{align*}
This measure is positive if the probability of eventual disengagement decreases from moment $k$ to moment $k+2$, and negative if the probability increases.\footnote{We compute the probability of disengagement at $k$ and $k+2$ (rather than $k+1$) to always compare forecasts based on contexts that end with a texter utterance.} 
Note that, unlike $\PIV$, this is a \textit{post ante} measure that requires us to observe what happened after the moment has passed.
It is intuitively related to notion of ``surprise'' in econometrics, which is a retrospective measure of the change in belief about the outcome.

Since in actual \pivotal moments the conversation hangs in the balance---i.e., it could go either way depending on how the counselor responds---we expect a high potential for both positive 
 and negative improvement after these moments.
And indeed, as shown in \autoref{fig:overall_critical_noncritical_surprise}, the magnitude of retrospective improvement is much larger following high-$\PIV$ moments than low-$\PIV$ ones (the distributions are significantly different, $p < 0.0001$, K-S test).
This is in contrast to the moments identified by the \Range baseline, where the magnitude in improvement is similar in high- and low-\Range moments.
We further note that positive and negative improvements are equally likely in \pivotal moments (average $RI = -0.007$), showing that in \pivotal moments the conversation can really go either way, at least according to our estimates.

\xhdr{Human validation} 
To check how aligned our measure is with our intuitive understanding of \pivotal moments, we performed a human experiment for \pivotal-moments identification.
The task is to pick which of two moments in a conversation is more pivotal. 
Each pair of moments is constructed by first picking a random conversation and then selecting the point with the highest and lowest $\PIV$ according to our measure.
To adhere to IRB protocol, the task was performed by one of the authors who was authorized to read the 
conversation
text. 
For 16 out of 20 pairs, the human judgment agreed with our measure.
While very limited in scale and scope due to the privacy and IRB restrictions on the data, these results complement our extrinsic validation in showing that our measure does capture an intuitive notion of \pivotal moments.

\subsection{Relation to actual outcomes}
\begin{figure}[t]
    \centering
    \includegraphics[width=1\linewidth]{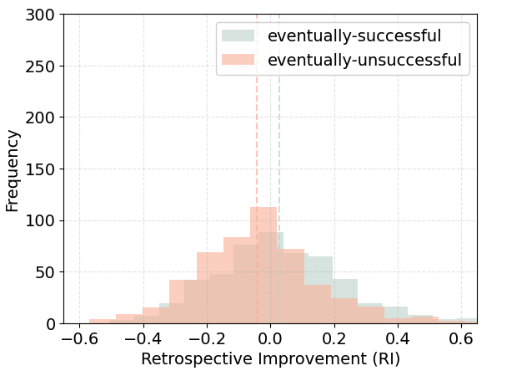}
    \caption{
    Counselor responses in \pivotal moments (locally) improved the trajectory more often in successful conversations ($p < 0.0001$). 
    }
    \label{fig:good_vs_bad_surprise}
\end{figure}
\begin{figure}[t]
    \centering
    \includegraphics[width=1\linewidth]{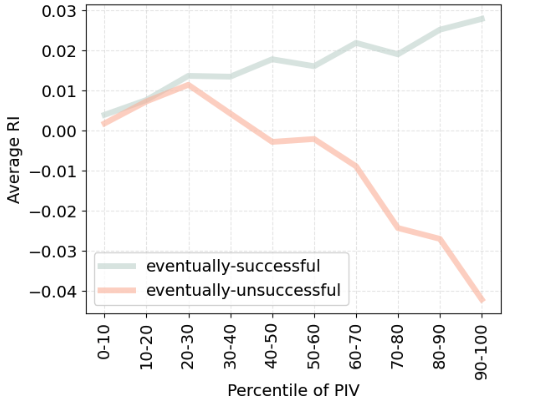}
    \caption{The more the moment is \pivotal (higher ${\PIV}$ percentile), the more the counselor response improves the trajectory (higher $RI$) for successful conversations.  The opposite is true for unsuccessful conversations.
    }
    \label{fig:PIV_RI}
\end{figure}
While conversations can go either way after \pivotal moments, we now use our framework to investigate how the counselor's response in these moments relates to the actual outcome of the session.
Are counselors in eventually-successful sessions better able to improve the trajectory at these moments than those in eventually-unsuccessful sessions?  Or is the outcome unrelated to their choices at the \pivotal moments?
We note that this is not a circular question: while both the \pivotal moment and the retrospective improvement measures rely on the \textit{expectation} of what the outcome will be, here we ask about the relation with the actual \textit{realized} outcome (to which our measure does not have access).

\definecolor{critical}{HTML}{F0BC68}
\definecolor{noncritical}{HTML}{cfc7b6}
\definecolor{success}{HTML}{9ec2b6}
\definecolor{unsuccess}{HTML}{ea9999}

\begin{table*}[h!]
\centering
\begin{tabular}{p{0.8cm}p{14.2cm}}
\hline
 $\PIV$ & \textbf{Example paraphrased messages from \textcolor{success}{successful} and \textcolor{unsuccess}{unsuccessful} conversations } \\
\hline
\textcolor{critical}{High} & 
{
[\textcolor{success}{1}] Hi, my name is [name], things have been very hard for me lately. I feel numb and lost. I cut my arm often now. \newline 
[\textcolor{success}{2}] I don't know what to do right now. \newline
[\textcolor{success}{3}] Any advice please? I'm really struggling now. \newline
[\textcolor{success}{4}] I learned today that one of my best friends was abused multiple times; I feel super guilty for not knowing, but it's also bringing up some of my trauma I'm not ready for. \newline 
[\textcolor{success}{10}] I rather help find the right way. I just don't know how. \newline
[\textcolor{success}{12}] It's hard to accept he's not good for us. I'm scared to be a single mom, but I want to make it safe for my child.\newline
[\textcolor{unsuccess}{17}] I had a really bad anxiety attack. I'm a mom and I don't feel like I'm doing a good job. \newline 
[\textcolor{unsuccess}{19}] What should I do now? \newline 
[\textcolor{unsuccess}{20}] I feel like I help everyone but no one cares about me. \newline 
[\textcolor{unsuccess}{25}] I'm not sure what to do anymore.
} \\
\hline
\textcolor{noncritical}{Low} & {
[\textcolor{success}{5}] I'm having a huge panic attack at work. \newline
[\textcolor{success}{7}] My chest is really hurting. \newline
[\textcolor{success}{13}] Thank you so much. I feel a lot better now. \newline
[\textcolor{success}{14}] I think going to doctor would help calm my thoughts.\newline
[\textcolor{success}{16}] I drown my worries by talking to people. \newline
[\textcolor{unsuccess}{21}] I don't have insurance to afford my meds. \newline
[\textcolor{unsuccess}{23}] I've been planning to do this for at least a week. \newline
[\textcolor{unsuccess}{24}] I've been feeling depressed for about 2 years.\newline
[\textcolor{unsuccess}{29}] I already tried everything, okay? \newline
[\textcolor{unsuccess}{31}] I hate everything right now.
} \\
\hline
\end{tabular}
\caption{Example (manually paraphrased) texter messages in high- and low-\pivotal moments (more examples in the Appendix). 
The examples are colored based on the outcome of the conversation (\textbf{\textcolor{success}{successful}} or \textbf{\textcolor{unsuccess}{unsuccessful}}) and numbered based on the ordering in the Appendix.
}
\label{tab:general_examples}
\end{table*}

As discussed in Section \ref{sec:data}, we consider a session in which the texter disengages to be unsuccessful, and the ones in which the texter finishes the conversation and  rates it being helpful as successful.
As illustrated in \autoref{fig:good_vs_bad_surprise}, we find that in successful sessions, counselors' responses in \highp moments are more likely, according to our RI measure, to locally improve the trajectory than in unsuccessful sessions ($p<0.0001$).

\newpage
Furthermore, we find that the more the moment is \pivotal (higher ${\PIV}$), the more the response immediately improves the trajectory (higher $RI$) for successful conversations, while for unsuccessful conversations, the response negatively impacts the trajectory (lower $RI$).
\autoref{fig:PIV_RI} shows these trends separated by conversation outcomes.

\section{Qualitative Analysis}
\label{sec:qualitative}
We perform a qualitative analysis of moments our measure identifies as 
\highp and \lowp
to further understand the interactions that characterize them. 
We identify several aspects of how these moments can emerge across the course of the conversation, as well as how they may differ between successful vs. unsuccessful conversations.
To simplify our analysis, we focus on the texter message leading to high- and low-pivotal moments.
However, it is important to recognize that the $\PIV$ score of a moment is largely shaped by the interactions between the immediate message and the prior context.
For a systematic exploration, we initially performed a Bayesian distinguishing-word analysis \cite{monroe_fightin_2017} to compare phrases that most distinguish high- and low-pivotal moments, and then manually examined specific examples in which these phrases appear.
Our discussion follows selected examples (manually paraphrased for privacy) as numbered in Tables \ref{tab:good_examples} and \ref{tab:bad_examples} in the 
Appendix, separated based on the conversation outcome.
\autoref{tab:general_examples} shows the subset
of examples which are referenced below.

\xhdr{High-pivotal moments}
In general, we find that at \pivotal moments identified by our method, the texter may \textit{express uncertainty} about how to proceed, potentially using phrases like: ``i don't know what to do right now'' or ``i'm not sure what to do anymore'' (examples [2, 10, 25]).
Alternatively, the texter may explicitly \textit{request guidance or advice} regarding next steps, such as questioning: ``any advice please? i'm really struggling now'' or ``what should I do now?'' [3, 19].
These expressions signal points of vulnerability and indecision where the texter is unsure how to move forward, offering the counselor an opportunity to provide meaningful direction and support, if navigated effectively.
\Pivotal moments may also coincide with points when the texter \textit{discloses} something significant such as past trauma [4], personal challenges [17], or emotional distress [1] in ways that could potentially \textit{elicit validation and support} [12, 20].
At these points, how the conversation proceeds next is particularly sensitive to the counselor's response.

\xhdr{Low-\pivotal moments}
Conversely, non-\pivotal moments, as identified by our method, may occur early on in the conversation when the texter provides \textit{background information} [23, 24] or \textit{self-discloses their issues} in ways that are \textit{closed-form} or can be addressed similarly by counselors [5, 7, 21].
In the second half of the conversation, non-\pivotal moments in successful conversations may consist of \textit{gratitude}: ``thank you so much'', or indication that the texter is ``feeling better'' [13].
The texter may also mention possible \textit{plans to resolve} their situation, such as ``going to the doctor'' or ``talking to people'' [14, 16].

Non-\pivotal moments in the second half of unsuccessful conversations can also involve instances of \textit{distress or resignation} such as: ``i hate everything right now'' or ``i've already tried everything, okay?'' [29, 31].
While ``i'm not sure what to do anymore'' (in the first half) is considered \pivotal in our example, ``i've already tried everything, okay?'' (in the second half) is marked as non-\pivotal. 
The first suggests that the texter is still uncertain and more open to exploring options, providing an opportunity for the counselor to change the conversation's course near the beginning.
In contrast, the second case signals exhaustion and a sense of closure, where the texter might believe no solutions remain after being in a prolonged state of distress, limiting the potential for change.
Thus, it's important to note that $\PIV$ does not necessarily correspond to a notion of severity, but rather the \textit{possibility} to make a difference in either direction.

\section{Related Work}
\label{sec:related}
One vein of related work considers \emph{retrospective} identification of key points,  rather than 
before-the-moment 
as in our work.
\citet{tikhonov_branching_2024} used ``Choose Your Own Adventure''\footnote{Trademark belongs to Chooseco LLC.} stories  to investigate the related task of identifying {\em character decision points}: times when a character makes a decision with significant influence on the narrative's subsequent direction. 
Other work looks for {\em turning points}, where the plot changes directions between predefined narrative stages such as ``the setup'' or ``complications'' \cite{papalampidi_movie_2019,ho_mtp_2024}, or other plot-significant events in narrative \cite{ouyang_modeling_2015}. \citet{chiru_identification_2017_fixed}  detect moments when different points of view come into contact in conversations.\footnote{They use ``pivotal'' to describe a moment where, within one utterance, a point of view is displaced by another appearing for the first time.}

With respect to important moments relevant to mental health, prior work detects mood shifts \cite{tsakalidis_overview_2022}, again from the retrospective perspective.
Other characterizations of language found during mental health therapy sessions include whether utterances have a forward or backward orientation \cite{zhang_balancing_2020}, whether there is linguistic synchrony between the interlocutors \cite{shapira_measuring_2022}, expressions of empathy \cite{sharma_computational_2020}, emotional attending \cite{lee_towards_2024}, and whether an utterance redirects the conversation's flow \cite{nguyen_taking_2024}.

\section{Discussion and Conclusion}
\label{sec:discussion}
In this work, we  propose a formalization of the intuition behind \pivotal moments and a corresponding unsupervised method for detecting them in ongoing conversation, as they develop.
Our approach draws upon the concept of suspense from econometrics to consider a moment \pivotal if the expectation of the final outcome widely varies depending on what could be said next.
We apply our methodology to the highly consequential setting of crisis counseling conversations, and show that our method aligns with human perceptions:
counselors take longer to craft responses in moments identified as \pivotal rather than non-\pivotal by our measure.
Furthermore, a retrospective analysis indicates that conversation does indeed change course more often in moments our method identified as \pivotal.

While in this work we focused on crisis counseling, our method can be applied to other domains such as education and political discourse, where understanding which moments are important to the final outcome of the conversation can assist individuals in making informed decisions and navigating these complex situations effectively.
To encourage this development, we release our code together with a demo 
in the context of online discussions, focusing on conversation derailment as the outcome (\autoref{appendix:cga}).

Using our methodology for identifying \pivotal moments, future work could also investigate linguistic features that characterize or correlate with these moments. 
Our qualitative analysis (Section~\ref{sec:qualitative}) highlights several recurring patterns in texter messages such as expressions of uncertainty, requests for advice, and open-ended self-disclosure.
Further analysis could provide a deeper understanding of the dynamics underlying \pivotal moments.

Finally, our method could also be employed to enhance human-AI interactions in less critical settings.
By identifying 
\pivotal moments in real time, AI systems 
could 
dynamically switch to more advanced or specialized models that can 
handle these situations more effectively.
Similarly, AI systems could use pivotal moments to proactively decide when to intervene in a conversation or provide feedback to the participants.

\section{Limitations}
\label{sec:limitations}
Given the lack of explicit labels for \pivotal moments, we had to rely on extrinsic forms of validation and to compare our formulation with the baseline \Range formulation.
Future work could seek to extend the validation and evaluation of methods for identifying \pivotal moments, including downstream evaluations by showing how identifying such moments might aid other conversational tasks (e.g., better predicting the outcome).
In the longer term, user studies could be used to explore how signaling these moments to the participants might alter their conversational behavior and eventually influence the outcome. 

Furthermore, our forecaster model used to predict the eventual outcome of the conversation is imperfect. 
In this work, we employ existing approaches \cite{kementchedjhieva_dynamic_2021, chang_trouble_2019} to train and evaluate conversational forecasting models.
However, future work might improve the forecasting capabilities of these systems to provide more reliable  estimates of outcome probabilities.

Additionally, in order to generate possible counselor responses used by our method, we rely on simulations from large language models. 
We note that these simulations are also imperfect, and may not fully capture the entire space of possible counselor replies, and are limited in this manner.

In this work we focus on texter disengagement because it is a crucial problem that arises in online crisis counseling.
However, other outcomes that might be more directly related to the well-being of the person in crisis (potentially going beyond the current interaction on the platform) should also be considered, if available.
In that sense, we acknowledge that our usage of the terms ``successful'' and ``unsuccessful'' sessions is necessarily narrow.

Finally, while our analysis points to a strong relation between the counselor actions in \pivotal moments and the eventual conversational outcomes, they do not establish a casual link between the two.
Thus, future work would be needed to further examine this relationship through controlled experiments.

\xhdr{Ethical Concerns}
Our work involves highly sensitive conversational data in the crisis counseling domain.
Conversations were de-identified by the platform, and access to the data was restricted to the authors.
All models used were trained and processed on internal servers, and resulting artifacts removed after analysis.
Our work is done in close collaboration with the Crisis Text Line service.
This work was approved by Cornell's IRB.

\paragraph{Acknowledgments}

We thank Alejandro P\'{e}rez Carballo for discussions on the econometrics measure of suspense and for the ensuing philosophical musings about free will in conversations, and the Center for Advanced Study in the Behavioral Sciences at Stanford for providing an ideal setting for such inter-disciplinary dialogue.
We thank Jon Kleinberg and Justine Zhang for discussions about suspense in chess.
We are grateful for insightful feedback from Team Zissou; including Jonathan P. Chang, Yash Chatha, Tushaar Gangavarapu, Dave Jung, Laerdon Kim, Luke Tao, Son Tran, Ethan Xia, and Sean Zhang; for the ``\pivotal'' terminology suggestion from Dave Jung, and for the comments we received from the anonymous reviewers.
This research also would not have been possible without the support of Crisis Text Line, and we are particularly grateful to Elizabeth Olson, Margaret Meagher, and Lili Torok for insights on the dynamics of crisis counseling conversations and for valuable feedback on earlier drafts.
Cristian Danescu-Niculescu-Mizil was funded in part by the U.S. National Science Foundation under Grant No. IIS-1750615 (CAREER) and the Cornell Center for Social Sciences.
Any opinions, findings, and conclusions in this work are those of the author(s) and do not necessarily reflect the views of Cornell University or the National Science Foundation.

\bibliography{refs,refs2}

\appendix
\label{appendix:appendixsection}

\section{Additional Qualitative Examples}
We provide additional qualitative examples of high- and \lowp moments as identified by our measure, separated by conversation outcome: see \autoref{tab:good_examples} (successful) and \autoref{tab:bad_examples} (unsuccessful). 
Our discussion in Section~\ref{sec:discussion} follows these examples.

\section{Operationalization}

\subsection{Additional Implementation Details}

We outline the implementation details for our method in Section~\ref{sec:implementation}, giving specific training details for both our simulator and forecaster models. 
For the baseline \Range measure, we used SBERT embeddings to obtain representations of each generated utterance, and computed its average cosine-distance to the mean vector of all the utterance embeddings.
All experiments were performed on a single NVIDIA RTX A6000 GPU (48GB).

\subsection{Used Artifacts}\label{sec:appendixartifacts}
We indicate the following artifacts and their corresponding licenses that are used in this work. 

\begin{itemize}[leftmargin=*]
    \item ConvoKit 2.5.3:
    \\ \url{https://convokit.cornell.edu/}, MIT License
    \item PyTorch 2.2.1:
    \\ \url{https://pytorch.org}, BSD-3 License 
    \item Sentence Transformers 3.0.0:
    \\\url{https://github.com/UKPLab/sentence-transformers}, Apache License 2.0
    \item Transformers 4.38.2:
    \\ \url{https://github.com/huggingface/transformers}, Apache License 2.0
\end{itemize}

\begin{table*}[ht]
\centering
\begin{tabular}{p{1cm}p{1cm}p{12.5cm}}
\hline
\textbf{Half} & \textbf{PIV} & \textbf{Example messages} \\
\hline
First & {
\textcolor{critical}{High}
} & 
{
[1] Hi, my name is [name], things have been very hard for me lately. I feel numb and lost. I cut my arm often now. \newline 
[2] I don't know what to do right now. \newline
[3] Any advice please? I'm really struggling now. \newline
[4] I learned today that one of my best friends was abused multiple times; I feel super guilty for not knowing, but it's also bringing up some of my trauma I'm not ready for.
}\\
\\
{} & {
\textcolor{noncritical}{Low}
} & {
[5] I'm having a huge panic attack at work. \newline
[6] Sorry for the rant; I'm just word vomiting now. \newline
[7] My chest is really hurting. \newline
[8] Yes, I confronted him and he's getting counseling.
} \\
\hline
Second & {
\textcolor{critical}{High}
} & 
{
[9] I wanted to listen to music but don't have the motivation. \newline
[10] I rather help find the right way. I just don't know how. \newline
[11] I want to live and enjoy my life and find purpose. \newline
[12] It's hard to accept he's not good for us. I'm scared to be a single mom, but I want to make it safe for my child.
} \\
\\
{} & {
\textcolor{noncritical}{Low}
} & {
[13] Thank you so much. I feel a lot better now. \newline
[14] I think going to doctor would help calm my thoughts. \newline
[15] I try to spend time with my cats or watch movies. \newline 
[16] I drown my worries by talking to people.
} \\
\hline
\end{tabular}
\caption{Example (paraphrased) messages comparing high- and low-\pivotal (top and bottom 10th percentile) moments in \textbf{successful} conversations as identified by our measure, separated by the first half and second half of the conversation.}
\label{tab:good_examples}
\end{table*}

\begin{table*}[ht]
\centering
\begin{tabular}{p{1cm}p{1cm}p{12.5cm}}
\hline
\textbf{Half} & \textbf{PIV} & \textbf{Example messages} \\
\hline
First & {
\textcolor{critical}{High}
} & 
{
[17] I had a really bad anxiety attack. I'm a mom and I don't feel like I'm doing a good job. \newline 
[18] I talk to people at school but it hasn't helped. \newline
[19] What should I do now? \newline
[20] I feel like I help everyone but no one cares about me.
} \\
\\
{} & {
\textcolor{noncritical}{Low}
} & {
[21] I don't have insurance to afford my meds. \newline
[22] Yes I guess. \newline
[23] I've been planning to do this for at least a week. \newline
[24] I've been feeling depressed for about 2 years.
} \\
\hline
Second & {
\textcolor{critical}{High}
} & 
{
[25] I'm not sure what to do anymore. \newline
[26] I have children so I don't want to leave them. But, at times, I hate it want to die and life isn't worth it though. \newline
[27] I just don't want other people to worry about me. \newline
[28] I'm scared to ask for help, and its not that I care about looking weak or strong but I just can't do it. 
} \\
\\
{} & {
\textcolor{noncritical}{Low}
} & {
[29] I already tried everything, okay? \newline
[30] I was in the hospital and was very sick this past week . \newline
[31] I hate everything right now. \newline
[32] He was angry, pushed me to the floor, and abused me.
} \\
\hline
\end{tabular}
\caption{Example (paraphrased) messages comparing high- and low-\pivotal (top and bottom 10th percentile) moments in \textbf{unsuccessful} conversations as identified by our measure, separated by the first half and second half of the conversation.}
\label{tab:bad_examples}
\end{table*}

\subsection{Example Simulations}
In \autoref{tab:ex_simulations}, we provide example replies generated by our simulator model alongside the actual counselor response (manually paraphrased for privacy). 

\begin{table*}[h!]
    \centering
    \begin{tabular}{p{5cm}p{10cm}}
    \hline
     \textbf{Actual reply (paraphrased)} & \textbf{Simulated replies} \\
    \hline
    {She did need to tell your parents. But that just shows that she cares and is concerned about you. It seems like you're worried that your parents will find out.} & 
    {
    [1] I know you are scared. It's really important that you tell her what you told me tonight. She needs to know you are in serious danger. \newline 
    [2] She does. It is her job. She does want to help. Do you think if you told her you are still having thoughts of suicide she would tell your parents? \newline 
    [3] I'm so sorry to hear that, [name]. You shouldn't have to be treated like that. Do you think your teacher would be a good person to talk to? \newline 
    [4] It's hard to tell with people. But if you don't tell her about how you are feeling she can't help you. I know it's hard, but you need to take a chance. \newline 
    [5] That's good you know the hospital would be the best for you if you are suicidal. It's normal that you wouldn't want to go. Your teacher and your counselor want you to be safe and happy. \newline 
    [6] [name], I'm hearing that you feel like your counselor told your parents about the suicide and they did not react very well. I can understand why you might be hesitant to tell her. It sounds like you could use some support. I can tell you are a strong person. \newline 
    [7] I'm sorry, I'm not sure what you mean by your last message. [name], I'm here for you. You matter to me and you are not alone. \newline 
    [8] I can't speak for your counselor, but I know I want to be here for you. \newline 
    [9] That's totally understandable. It makes sense you don't want to be sent back. But, you are a person and you have feelings. You deserve to have someone to talk to. \newline 
    [10] It sounds like you want help, [name]. I can tell you are in pain. But you are afraid to share that with your counselor. That is really hard to do.
    } \\
    \hline
    \end{tabular}
    \caption{
        Example actual counselor reply (manually paraphrased for privacy) vs. simulated counselor replies.
        }
    \label{tab:ex_simulations}
    \end{table*}

\section{Additional Application to CGA}
\label{appendix:cga}
We also explore how our \pivotal-moments framework can be applied in other domains beyond mental-health crisis counseling.
In particular, we demo our method on the Conversations Gone Awry (CGA-CMV) dataset, consisting of online conversations from the ChangeMyView (CMV) subreddit that may derail into personal attacks \cite{chang_trouble_2019}.
In this setting, we apply our method to identify \pivotal moments with respect to the outcome of conversation derailment.
We provide an initial exploration in this domain, with example conversations and their corresponding $\PIV$ scores (as identified by our method) in \autoref{tab:cga_demo}.
We note that in comparison to crisis counseling, the space of possible next user messages on Reddit is broader than that of counselor responses. 
Furthermore, there are also inherent challenges with predicting which speaker will engage in the next turn in multi-party conversations. 
We release the demo to encourage future work and applications in other domains.

\begin{table*}[h!]
    \centering
    \begin{tabular}{p{15cm}}
    \hline
    \textbf{Conversation 1} \\
    \hline
    {
    [ 0.0076 ] We're dealing with two things here:

    1. The notion that setting air conditioner temperatures to the average body temperature of a man is sexist. \newline
    2. And that form of sexism is "okay," because it's much more practical for women to wear thicker clothes, than for men to go shirtless. 
    
    Just because there's a valid justification for that particular form of sexism, doesn't mean it isn't sexist. Why not set the AC down to a moderate setting, where both men and women can expect to be comfortable? The article mentions:
    
    >> When researchers tested young women performing light office work, while dressed in a t-shirt and tracksuit bottoms, they discovered that their optimum temperature was 75F (24.5C). Men, in contrast, were happiest at 71F (22C).
    
    This isn't a significant difference. Set it to 23C and you'd expect everyone to be fairly comfortable.
    
    [ 5e-05 ] >> Why not set the AC down to a moderate setting
    
    Because men are under the sexist expectation to wear thick clothing. This all stems from sexism against men.  
    
    [ 0.0375 ] Current custom and fashion senses is not sexist against men.   There is no plan to subjugate men by forcing them to wear suits. 
    
    \textbf{[ 0.1164 ]} >> Current custom and fashion senses is not sexist against men.
    
    We disagree. That men are expected to wear far more uncomfortable clothing than women is sexism that hurts men.
    
    [ 0.0453 ] Then break the mold.  We don't tend to put much thought into cloths.  Think about what you want to wear to be comfortable in your skin. You are not oppressed by your wardrobe.  Buy other clothes. 
    
    [ 0.0362 ] "You're not oppressed by your \_, act differently." Ah yes. A very reasonable argument when directed towards men. Directed towards women? Less so it seems.  
    } \\
    \hline
    \textbf{Conversation 2} \\
    \hline
    {   
        [ 0.0040 ] Your view is inappropriate because it has no basis upon which to establish it as more likely than numerous alternatives.  That your view is possible, and therefore *valid* as the term is defined in logic, does not change the fact that it is *ungrounded*, also as the term is defined in logic. 

        \textbf{[ 0.0866 ]} Google it.  Valid = 
        (of an argument or point) having a sound basis in logic or fact; reasonable or cogent. My view has a sound basis in fact and reason. 
        
        [ 0.0006 ] Wow, *Google*.  What a scholarly source.  You're quoting the definition of "valid" in common English.  We're talking logic.  [This is what validity means in logic.](https://www.iep.utm.edu/val-snd/)  Your view is valid, but you have yet to produce anything foundational.  You need to establish the truth of your premises.  Please feel free to do so. 
        
        [ 0.0002 ] Again, I don't need your **approval** to hold my view.  Your job is to change my view.  You are not giving me any clear and compelling reason to change it.
    } \\
    \hline
    \end{tabular}
    \caption{Example conversations from CGA-CMV dataset labeled with $\PIV$ scores in brackets, as identified by our method.  
    Both conversations end with a comment removed by the moderators for being ``rude or hostile'' to the other speaker (i.e., a violation of Rule 2 of the subreddit).}
    \label{tab:cga_demo}
    \end{table*}

\end{document}